\documentclass[11pt]{article}

\usepackage[]{EMNLP2022}

\usepackage{times}
\usepackage{latexsym}

\usepackage[T1]{fontenc}

\usepackage[utf8]{inputenc}

\usepackage{microtype}

\usepackage{tabularx,booktabs, multirow, graphicx, comment}
\usepackage{hyperref, url}

\begin{document}

\title{Denoising-based UNMT is more robust to word-order divergence than MASS-based UNMT}

 \author{Tamali Banerjee \\ IIT Bombay, India \\ tamali@cse.iitb.ac.in
         \AND
         Rudra Murthy V \\ IBM Research Lab, India \\ rmurthyv@in.ibm.com 
\And
         Pushpak Bhattacharyya \\ IIT Bombay, India \\  pb@cse.iitb.ac.in  }


\maketitle

\begin{abstract}


We aim to investigate whether UNMT approaches with self-supervised pre-training are robust to word-order divergence between language pairs or not. We achieve this by comparing two models pre-trained with the same self-supervised pre-training objective. The first model is trained on language pairs with different word-orders, and the second model is trained on the same language pairs with source language re-ordered to match the word-order of the target language. 
Ideally, UNMT approaches which are robust to word-order divergence should exhibit no visible performance difference between the two configurations. 
In this paper, we investigate two such self-supervised pre-training based UNMT approaches, namely Masked Sequence-to-Sequence Pre-Training, (MASS) (which does not have shuffling noise) and Denoising AutoEncoder (DAE), (which has shuffling noise).

We experiment with five English$\rightarrow$Indic language pairs (\textit{i.e.} en-hi, en-bn, en-gu, en-kn, and en-ta) where word-order of the source language is SVO (Subject-Verb-Object) and the word-order of the target languages are SOV (Subject-Object-Verb). 
We observed that for these language pairs, DAE-based UNMT approach consistently outperforms MASS in terms of translation accuracies.
Moreover, bridging word-order gap using reordering improves the translation accuracy of MASS-based UNMT models while it cannot improve the translation accuracy of DAE-based UNMT models. 
This observation indicates that DAE-based UNMT is more robust to word-order divergence than MASS-based UNMT. 
Word-shuffling noise in DAE approach could be the possible reason for the approach being robust to word-order divergence. 
\end{abstract}

\section{Introduction}

Recently, self-supervised pre-training based approaches to Unsupervised Neural Machine Translation (UNMT) \citep{conneau2019cross,song2019mass,liu-etal-2020-multilingual-denoising} have become the norm. These UNMT models are trained in two stages \textit{viz} (i) self-supervised pre-training and (ii) iterative back-translation. The self-supervised pre-training phase typically involves training an encoder-decoder transformer on the combined monolingual corpora of two or more languages using self-supervision. The second stage employs iterative back-translation to train the model for translation in all directions.

The approaches differ in the self-supervised objective employed during the pre-training phase. For instance, \citet{conneau2019cross} employ Masked Language Modeling (MLM) objective to train an encoder. Here, some of the tokens in the input sequence are masked and the objective is to predict the masked tokens.  MASS \citep{song2019mass} employs the masking of consecutive tokens and employs a decoder to generate the masked tokens. 
 \citet{liu-etal-2020-multilingual-denoising} use Denoising Auto-Encoder (DAE) to pre-train the model. The noise functions considered are (a) removing spans of text and replacing them with a mask token and (b) sentence shuffling. 

While the above approaches show promising results for closely-related language-pairs, however, language divergence between source and target languages can adversely impact the benefits obtained from the joint-training, resulting in poor translation by UNMT models. 

Language divergence can be observed between two languages at various stages of NLP \citep{bhattacharyya2012natural}. Recently, several approaches have been proposed to address the issue of lexical divergence between languages \cite{chronopoulou-etal-2021-improving,banerjee-etal-2021-crosslingual, khatri2021simple}. 
However, the impact of word-order divergence on UNMT is still not explored. Previously, \citet{sun2021unsupervised} experimented with the iterative UNMT approach proposed by  \citet{lample-etal-2018-phrase} which does not employ self-supervised pre-training and is found to be sensitive to word-order divergence.

Recently, \citet{abdou-etal-2022-word} showed that shuffled language models such as DAE can still surprisingly exhibit information about naturalistic word order. 

Re-ordering (synonymous with pre-ordering) has been used as a tool in Machine Translation to bridge the word-order gap among languages. While \citet{du2017pre} found re-ordering redundant for NMT systems, \citet{zhao-etal-2018-exploiting} observed improvement from re-ordering by reducing the number of under-translation cases. In a transfer learning scenario, \citet{murthy2018addressing} observed improvement in result by re-ordering the assisting source language sentences to match the word-order of the actual source language before training the parent model.

While the use of re-ordering tools sounds promising, re-ordering tools require either an (i) word re-ordered data to train a supervised re-ordering tool or ii) parser and re-ordering rules to use a rule-based system. Also, 
it is unclear if existing UNMT approaches are robust to word-order divergence or not.

In this paper, we aim to investigate the effect of word-order divergence in state-of-the-art UNMT systems. In this direction, we compare MASS \citep{song2019mass} and DAE \citep{artetxe2018unsupervised} on language pairs with different word-orders.  We experiment with five language pairs in which the word-order of the source language does not match the word-order of the target language. For each language-pair, we compare (i) UNMT model trained on the source$\rightarrow$target language data (original data), with (ii) UNMT model trained on the re-ordered source $\rightarrow$ target data (re-ordered data). Here, \textit{re-ordered source} refers to the source sentences re-ordered to match the word-order of the target language. In the former case, word-order divergence is present while in the latter case word-order divergence is bridged. 

We observe from our experiments that there is no significant difference in terms of translation accuracy between DAE-based models trained on original data and re-ordered data. It signifies that DAE-based UNMT is robust to the word-order difference between the two languages compared to MASS-based UNMT. We also observe that in most language pairs, the UNMT model trained using DAE approach with original data produces better quality translations compared to other models.



 
\section{Approaches used}
We use MASS-static and DAE-static approaches proposed by \citet{banerjee-etal-2021-crosslingual} to handle lexical divergence. For both approaches, the embedding layers of the UNMT model are initialised with unsupervised cross-lingual embeddings and though out the training (pre-training followed by fine-tuning) embedding layers are not updated (static). For every token, the input to the model is the summation of cross-lingual word embedding and positional encoding. Unlike the original MASS approach, language code is not added here.

\subsection{MASS objective}
In MASS objective \citep{song2019mass}, an n-gram tokens of size \textit{k} (where \textit{k} is the half of the length of the sentence) in the input sentence are chosen randomly. Within that fragment, 80\% tokens are masked, 10\% tokens are replaced by random tokens, and the remaining 10\% tokens are kept unchanged. Then, the model is trained to generate the missing n-grams. 

\subsection{DAE objective}
Here, we employ DAE-based pre-training, specifically, we add random noise to the input sentence before giving it as input and the model is trained to generate the entire original sentence. While any noise could be employed \citep{liu-etal-2020-multilingual-denoising} we use the word shuffle, word mask, and word deletion noise introduced by \citet{artetxe2018unsupervised}. Here, shuffling means swapping neighboring words which results in local shuffling in a sentence.

\begin{table*}[!t]
\centering
    \begin{minipage}{0.45\textwidth}
    \centering
    \begin{tabular}{l r }
            \toprule
            \textbf{Language} & \textbf{\# train sentences}\\
            \midrule
            English (en) & 54.3 M\\
            \midrule
            Hindi  (hi) & 63.1 M \\
            Bengali  (bn) & 39.9 M \\
            Gujarati  (gu) & 41.1 M \\
            Kannada (kn) & 53.3 M \\
            Tamil (ta) & 31.5 M \\
            \bottomrule
    \end{tabular}
    \end{minipage}\hfill
    \begin{minipage}{0.45\textwidth}
    \centering
    \begin{tabular}{l  c c}
            \toprule
            \textbf{Language-pair} & \textbf{\# valid} & \textbf{\# test} \\
            & \multicolumn{1}{c}{\textbf{ sentences}} & \multicolumn{1}{c}{\textbf{ sentences}}\\
            \midrule
             en - hi & 1000 & 2390  \\
             en - bn & 1000 & 2390 \\
             en - gu & 1000 & 2390 \\
             en - kn & 1000 & 2390 \\
             en - ta & 1000 & 2390 \\
            \bottomrule
    \end{tabular}
    
    \end{minipage}
    \caption{Dataset statistics before noise removal}
    \label{tab:dataset details 1}
\end{table*}

\begin{table*}[!t]
\centering
    \begin{minipage}{0.45\textwidth}
    \centering
    \begin{tabular}{l r }
            \toprule
            \textbf{Language} & \textbf{\# train sentences}\\
            \midrule
            English (en) & 52.2 M\\ 
            Reordered English (en) & 52.2 M\\
            \bottomrule
    \end{tabular}
    \end{minipage}\hfill
    \begin{minipage}{0.45\textwidth}
    \centering
    \begin{tabular}{l  c c}
            \toprule
            \textbf{Language-pair} & \textbf{\# valid} & \textbf{\# test} \\
            & \multicolumn{1}{c}{\textbf{ sentences}} & \multicolumn{1}{c}{\textbf{ sentences}}\\
            \midrule
             en - hi & 711 & 1781  \\
             en - bn & 711 & 1781 \\
             en - gu & 711 & 1781 \\
             en - kn & 711 & 1781 \\
             en - ta & 711 & 1781 \\
            \bottomrule
    \end{tabular}
    
    \end{minipage}
    \caption{Dataset statistics after noise removal}
    \label{tab:dataset details 2}
\end{table*}

\subsection{UNMT with re-ordering} 
To train a UNMT model with re-ordered data, we first re-order source sentences of monolingual data to match the word-order of the target language.
UNMT model is then trained with the re-ordered-source and target monolingual data.
To test the model in source$\rightarrow$target direction, source test sentences are re-ordered before giving it as input to the model. We do not use this model as a target$\rightarrow$source system, as it generates source sentences in re-ordered form.

\section{Experimental setup}
\label{sec:setup}

We have four sets of UNMT models in out experiment. Two of them are trained using MASS-static and other two are trained using DAE-static. For each of these UNMT approaches, we train one model on original data, another one model on re-ordered data.

\subsection{Language and datasets}
We use monolingual data of 6 languages \textit{i.e.} English (en), Hindi (hi),  Bengali (bn), Gujarati (gu), Kannada (kn) and Tamil (ta). Here, except English, all other languages are Indian languages having Subject-Object-Verb (SOV) word-order. For English which is from the European language family, the word-order is Subject-Verb-Object (SVO). For all language pairs used in our experiment, we keep English as source and one of these 5 Indic languages as target resulting in 5 language pairs, \textit{i.e.} en-hi, en-bn, en-gu, en-kn, and en-ta.

We use monolingual data provided by IndicCorp dataset \cite{kakwani-etal-2020-indicnlpsuite} as training data. We use English-Indic validation and test data provided in WAT 2021 Shared task  \citep{nakazawa-etal-2021-overview}. We present the data statistics of the datasets in Table \ref{tab:dataset details 1}. 

\begin{table*}[!htb]
    \centering
    \resizebox{\textwidth}{!}{ 
    \begin{tabular}{ll  ll ll ll}
    \toprule
     \textbf{Evaluation metrics} & \textbf{UNMT approaches} &  \multicolumn{6}{c}{\textbf{Translation accuracies on different language-pairs} } \\
    \cmidrule(lr){1-1} \cmidrule(lr){2-2} \cmidrule(lr){3-8}
     \multirow{2}{*}{ } & \multirow{2}{*}{ } & \multicolumn{2}{c}{en - hi}   & \multicolumn{2}{c}{en - bn} & \multicolumn{2}{c}{en - gu} \\ \cmidrule(lr){3-4} \cmidrule(lr){5-6} \cmidrule(lr){7-8} 
      &  & S $\rightarrow$ T & T  $\rightarrow$ S & S $\rightarrow$ T & T  $\rightarrow$ S & S $\rightarrow$ T & T  $\rightarrow$ S \\  
    \midrule
     \multirow{4}{*}{BLEU} & MASS-static & 14.16 & 14.03 & 1.51 & 2.77 & 5.31 & 6.25 \\
     & MASS-static + R &  14.63 ($\uparrow$ 0.47) & - & 3.04 ($\uparrow$ 1.53) & - & 8.62 ($\uparrow$ 3.31) & - \\
    \cmidrule{2-8}
    & DAE-static &  \textbf{21.03} & \textbf{21.89} & 2.88 & \textbf{4.39} & \textbf{10.60} & \textbf{14.78}  \\
    & DAE-static + R & 15.22   ($\downarrow$5.81) & - & \textbf{3.27} ($\uparrow$ 0.39) & - & 8.72 ($\downarrow$ 1.88) & -  \\
    \midrule
    \multirow{4}{*}{CHRF} & MASS-static &  39.45 & 46.07 & 25.39 & 29.56 & 31.08 & 36.87 \\
     & MASS-static + R & 41.77 ($\uparrow$ 2.32) & - & 28.85 ($\uparrow$ 3.46) & - & 37.45 ($\uparrow$ 6.37) & - \\
    \cmidrule{2-8}
    & DAE-static &  \textbf{45.63} & \textbf{52.21} & 28.36 & \textbf{34.61} & \textbf{38.55} & \textbf{45.74} \\
    & DAE-static + R &  42.18 ($\downarrow$ 3.45) & - & \textbf{29.65} ($\uparrow$ 1.29) & - & 37.58 ($\downarrow$ 0.97) & - \\ 
    \bottomrule 
    \\
    \bottomrule 
      &  & \multicolumn{2}{ c}{en - kn}  & \multicolumn{2}{ c }{en - ta} \\
     \cmidrule(lr){3-4} \cmidrule(lr){5-6} 
      &  
    & S $\rightarrow$ T & T  $\rightarrow$ S & S $\rightarrow$ T & T  $\rightarrow$ S \\  
    \midrule
     \multirow{4}{*}{BLEU} & MASS-static & 3.08 & 5.11 & 1.81 & 2.70  \\
     & MASS-static + R & \textbf{4.48} ($\uparrow$ 1.4) & - & 2.57 ($\uparrow$ 0.76) & - \\
    \cmidrule{2-8}
    & DAE-static & 4.42 & \textbf{9.40} & 2.52 & \textbf{4.77} \\
    & DAE-static + R & 4.18 ($\downarrow$ 0.24) & - & \textbf{2.65} ($\uparrow$ 0.13) & -  \\
    \midrule
    \multirow{4}{*}{CHRF} & MASS-static & 30.73 & 33.83 & 31.40 & 28.73 \\
     & MASS-static + R & \textbf{36.47} ($\uparrow$ 5.74) & - & 34.34 ($\uparrow$ 2.94) & - \\
    \cmidrule{2-8}
    & DAE-static & 35.89 & \textbf{40.33} & 33.00 & \textbf{34.48} \\
    & DAE-static + R & 36.46 ($\uparrow$ 0.57) & - & \textbf{35.52} ($\uparrow$ 2.52) & - \\
    \bottomrule 
    \end{tabular}
    }
    \caption{Translation accuracies of MASS-pretrained and DAE-pretained UNMT models with/without re-ordering (R) for language pairs in both directions (S and T refer to source and target). The values marked in bold indicate the best scores for each language pair. The values within parenthesis signify improvement/degradation in BLEU/CHRF of the model from the model above them.}
    \label{result}
\end{table*}

\begin{figure*}[!htb]
    \centering
    \includegraphics[width=0.75\textwidth]{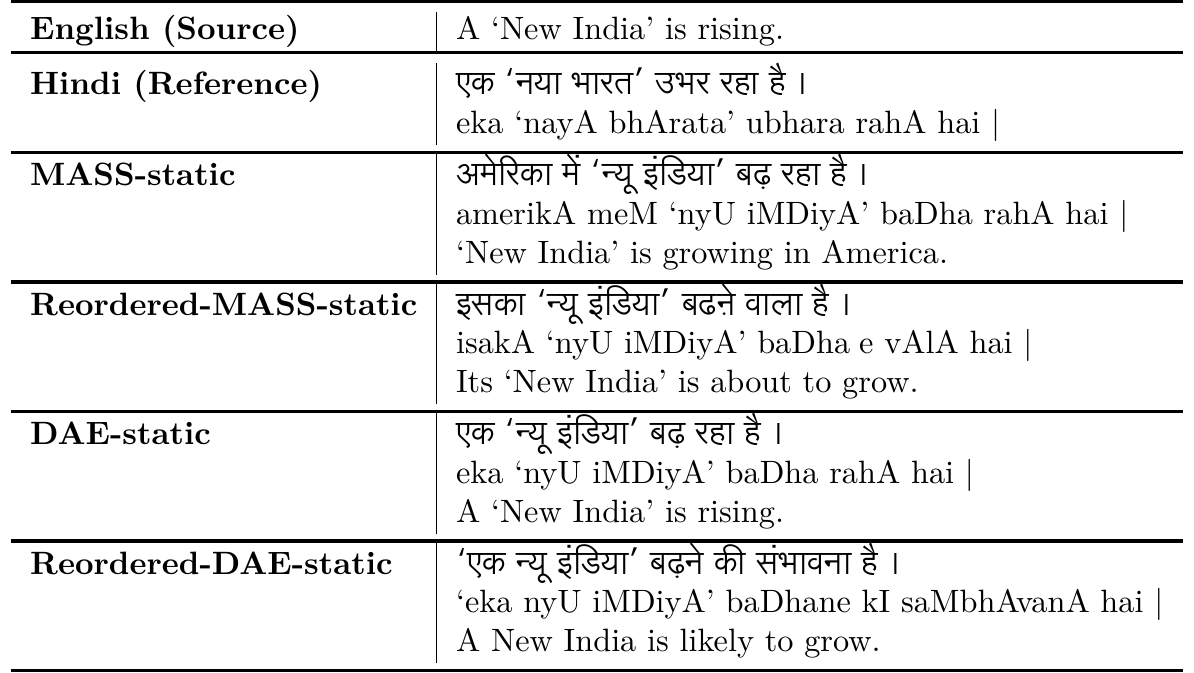}
    \caption{An example of en$\rightarrow$hi translations using different UNMT models}
    \label{fig:example2}
\end{figure*}

\begin{figure*}[ht!]
    \centering
    \includegraphics[width=\textwidth]{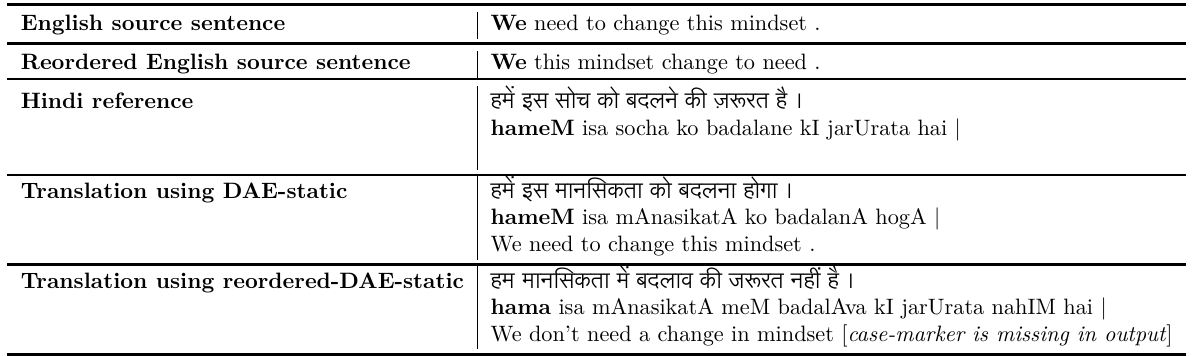}
    \caption{Translation example where re-ordering creates ambiguity}
    \label{fig:dativ}
\end{figure*}

\begin{figure*}[!htb]
    \centering
    \includegraphics[width=0.7\textwidth]{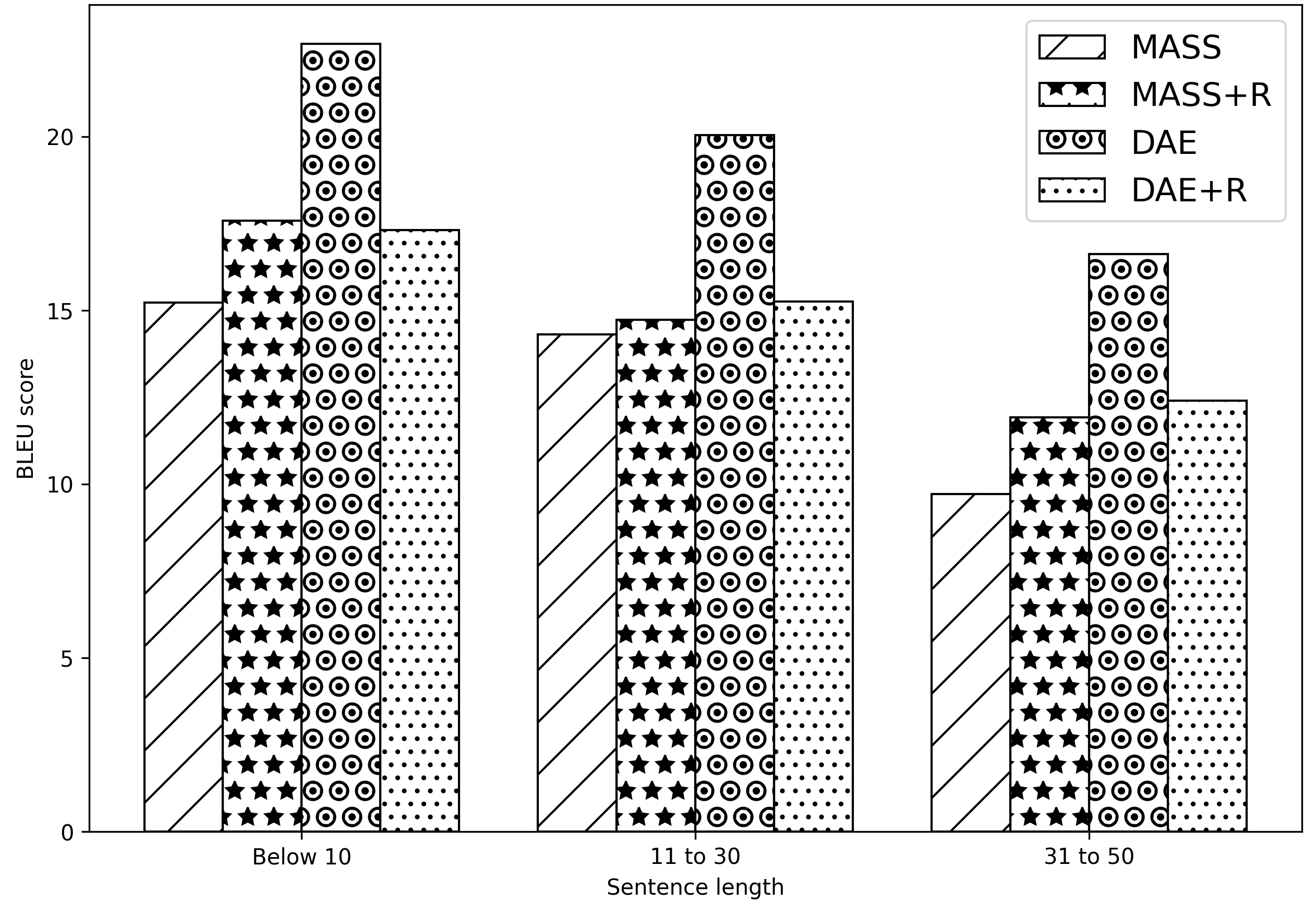}
    \caption{BLEU score distribution w.r.t. sentence length. Language pair: en$\rightarrow$hi.}
    \label{fig:sentbleuplot}
\end{figure*}

\begin{figure*}[ht!]
    \centering
    \includegraphics[width=\textwidth]{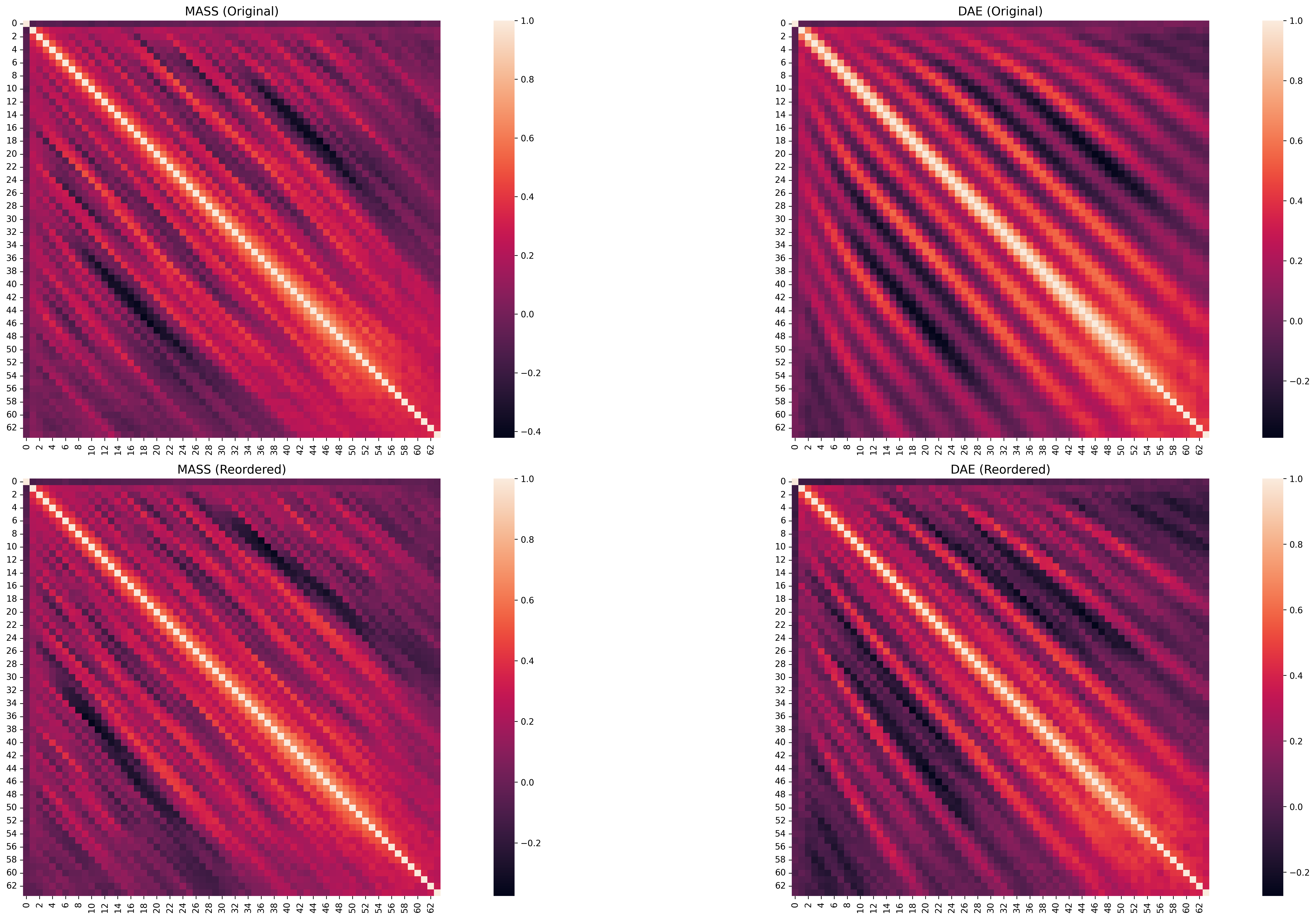}
    \caption{Comparison among the position embeddings of four UNMT models (\textit{e.g.} MASS-original, DAE-original, MASS-reordered and DAE-reordered). Each heatmap shows position-wise cosine similarity between the position embeddings of a UNMT model. Lighter color in the figures denotes the higher similarity. Language pair: en$\rightarrow$hi.}
    \label{fig:posemb}
\end{figure*}

\subsection{Preprocessing tools}

We have tokenised the English corpus using \textit{Moses} \citep{koehn2007moses} and the Indic corpora using \textit{Indic NLP Library} \citep{kunchukuttan2020indicnlp}. 

We use Generic rules of CFILT-pre-order  \citep{chatterjee2014supertag} for re-ordering English sentences to match word-order of Indic languages in  training, validation and test data.

For BPE segmentation \citep{sennrich2016improving}, we use \textit{FastBPE}\footnote{\url{https://github.com/glample/fastBPE}} jointly on the source and target data with the number of merge operations set to 100k. 

\subsection{Crosslingual embeddings}
We combine the BPE-segmented monolingual corpora of source and target to train the embeddings using skip-gram model of \textit{Fasttext}  \citep{bojanowski2017enriching}. From the embeddings of combined vocabulary we get the embeddings of the language-specific vocabularies. To improve the quality of crosslingual embeddings of the two languages, we apply \textit{Vecmap} \citep{artetxe2018robust} which we used in our experiments.
 
\subsection{Reordering noise removal}
The reordering tool generates parse errors when it fails to reorder a sentence correctly. To make a fair comparison, we removed all those monolingual sentences (for which the reordering tool generated parse error) from both original and re-ordered data. We did the same (removed their translations too) for valid and test parallel data. It made our data size shrink in a small amount. We perform all the experiments on the remaining data. Details of the remaining data are in Table \ref{tab:dataset details 2}.

\subsection{Network and evaluation}

We use MASS code-base \cite{song2019mass} to run our experiments. The model is trained using an epoch size of $0.2M$ steps and a batch size of 64 sentences (token per batch $3K$)). For each of pretraining and finetuning steps, we train the models for 50 epochs maximum. However, we stop the training if the model converges before the max-epoch is reached based on validation split loss.  For MASS pretraining, we use word-mass of $0.5$ \textit{i.e.,} $50\%$ of the tokens in the sentence are masked. Other parameters are default parameters given in the code-base. For DAE pretraining, we use word-shuffle $3$,  word-dropout $0.1$, and word-blank $0.1$.

We report the performance of the translation system using BLEU \citep{papineni2002bleu} and CHRF \citep{popovic2016chrf} scores using SacreBLEU \citep{post-2018-call}. CHRF (Character n-gram F-score) is beneficial for the evaluation of morphologically rich languages \textit{e.g.} Indic languages.

\begin{figure}[!htb]
    \centering
    \includegraphics[width=0.5\textwidth]{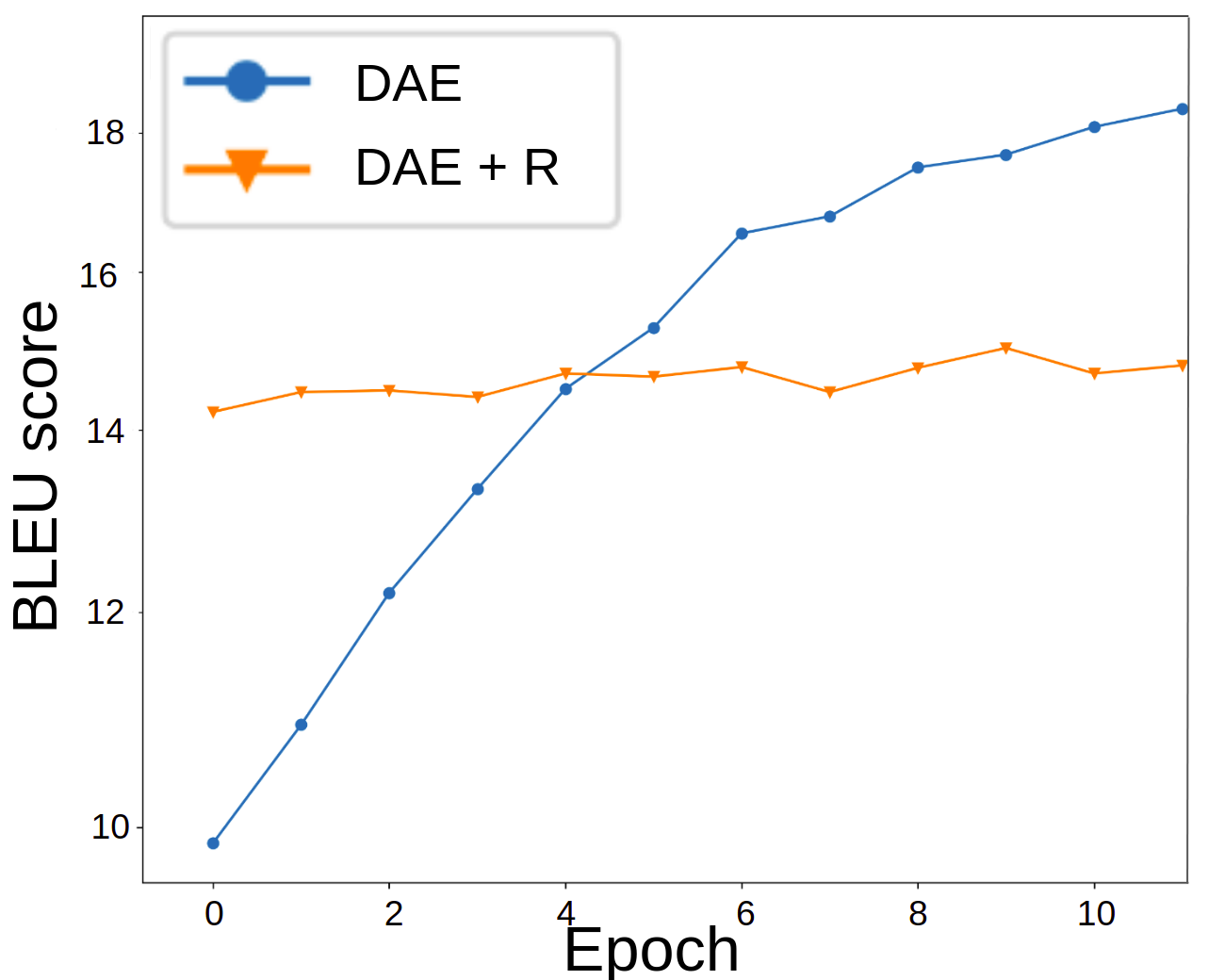}
    \caption{Change in BLEU score of translations by DAE-static UNMT models with increasing number of epochs during finetuning from epoch 0 to 10. Comparison between the model trained with original data and the model trained with re-ordered data. Language pair: en$\rightarrow$hi. }
    \label{fig:bleuplot}
\end{figure}

\section{Result and analysis}

The BLEU and CHRF scores of the UNMT models with or without re-ordering (R) from our experiments are as shown in Table \ref{result}. We also report how much improvement or degradation the incorporation of re-ordering caused to the translation quality of the UNMT model in terms of BLEU and CHRF scores. We do not show the result in target$\rightarrow$source direction for re-ordered UNMT models. This is because given a target sentence as input these re-ordered models generate its translation in source language but in re-ordered form. For most of the language pairs, the UNMT model which is trained on DAE-static approach with original data produced the best results. Our manual analysis also confirms the fact as shown in Figure \ref{fig:example2} with an example. Here, translation by DAE-static model is similar to reference sentence, while in  translations by other models are not adequate.

For MASS-static models, the models which are trained on re-ordered data surpass the models which are trained on original data. This indicates that MASS-static UNMT is not robust to word-order divergence. Intuitively, the reason lies in the self-supervised pre-training objective function which does not use shuffling in it. As a result, it makes the MASS model rigid about remembering position information.

A different scenario is observed in case of DAE-static models. Here, models give the best results even without re-ordering. It indicates that DAE-static is robust to word-order divergence. Presumably, presence of shuffling in the DAE objective function makes the model flexible about remembering/forgetting position information.

Surprisingly, for most of the language-pairs, training the model with re-ordered data degrades the BLEU and CHRF scores of output. This is due to the ambiguity caused by re-ordering an SVO language in SOV without case-markers. Here, we show such an example in Figure \ref{fig:dativ}. In the example shown in Figure \ref{fig:dativ}, reference translation of the subject `we' is `hameM' which is in dative case. When we re-order the English sentence, it remains `we' which is of subjective case and whose frequent Hindi translation is `hama'. Consequently, in the translation produced by re-ordered model, we see the word `hama' instead of the desired word `hameM'.


Figure \ref{fig:sentbleuplot} reveals that MASS is more sensitive to word-order divergence for short and long sentences. DAE-static model significantly outperforms other approaches across all sentence lengths.

Figure \ref{fig:bleuplot} shows that when DAE is trained with re-ordered data, the finetuning model starts generating good quality translation with the first epoch itself. However, it cannot improve more and after some epochs model trained with original data surpasses it.

Figure \ref{fig:posemb} shows the visualization of position-wise cosine similarity of each position embedding. The point at (\textit{i, j}) indicates the similarity between the $i^{th}$ position and the $j^{th}$ position. The graph shows the visualization for both DAE and MASS pre-trained UNMT models on the data with and without re-ordering. We observe that in MASS, the position embeddings are similar to itself and words in a local neighborhood of 2 or 3 positions. In DAE, the position embeddings are similar to a larger local neighborhood which could be attributed to the local shuffling noise. We also observe that, for DAE models, the position embeddings are similar to a larger number of neighbouring positions in presence of word-order divergence (\textit{i.e.} without re-ordering) compared to the number of neighbouring position having similar position embedding in presence of word-order similarity (\textit{i.e.} with re-ordering).


\section{Conclusion}
This paper investigates the effect of word-order divergence on MASS-based and DAE-based UNMT. To test that, we check whether re-ordering source data to match the word-order of target data before training improves the translation accuracy of UNMT models or not. Our results indicate that DAE-based UNMT is more robust to word-order divergence than MASS-based UNMT. Additionally, DAE-based UNMT model has the added advantage of having a single UNMT model in both directions (English to Indic and Indic to English languages) whereas the model trained on re-ordered data could be only used to translate in one direction (reordered English $\rightarrow$ Indic) only.

 
\bibliography{anthology, custom}
\bibliographystyle{acl_natbib}

\end{document}